# An Empirical Comparison of Forgetting Mechanisms for UCB-based Algorithms on a Data-Driven Simulation Platform

Minxin Chen
*Swjtu-Leeds Joint School, South West Jiao Tong University, Chengdu, China*
*2022116019@my.swjtu.edu.cn*

Abstract: Many real-world bandit problems involve non-stationary reward distributions, where the optimal decision may shift due to evolving environments. However, the performance of some typical Multi-Armed Bandit (MAB) models such as Upper Confidence Bound (UCB) algorithms degrades significantly in non-stationary environments where reward distributions change over time. To address this limitation, this paper introduces and evaluates FDSW-UCB, a novel dual-view algorithm that integrates a discount-based long-term perspective with a sliding-window-based short-term view. A data-driven semi-synthetic simulation platform, built upon the MovieLens-1M and Open Bandit datasets, is developed to test algorithm adaptability under abrupt and gradual drift scenarios. Experimental results demonstrate that a well-configured sliding-window mechanism (SW-UCB) is robust, while the widely used discounting method (D-UCB) suffers from a fundamental learning failure, leading to linear regret. Crucially, the proposed FDSW-UCB, when employing an optimistic aggregation strategy, achieves superior performance in dynamic settings, highlighting that the ensemble strategy itself is a decisive factor for success.

## 1 INTRODUCTION

In systems such as personalized recommendations or online advertising (Cui et al., 2020; Tan et al., 2025), platforms must make continuous decisions in real-time interactions with users to maximize long-term objectives like click-through rates or commercial revenue. The Multi-Armed Bandit (MAB) model provides a solid theoretical framework for such sequential decision problems by dynamically balancing "exploration" and "exploitation" (Lattimore & Szepesvári, 2020). In the standard MAB paradigm, an agent selects one option from a set of candidates ("arms") in each round and observes its stochastic reward. The core objective is to minimize cumulative regret in an environment where the reward distributions are unknown and potentially changing. This characteristic of adaptive learning has led to the widespread deployment of various MAB algorithms in modern, large-scale recommendation and advertising platforms.

Classical MAB algorithms, such as the Upper Confidence Bound (UCB) or Thompson Sampling (TS), are proven to be theoretically optimal in stationary environments where reward distributions are fixed. However, in real-world applications, environmental non-stationarity is the norm rather than the exception. User interests evolve, and the popularity of content fluctuates due to seasonality or sudden events, leading directly to drift in the reward distributions. Under such dynamic conditions, the performance of traditional algorithms degrades significantly, as they may persistently select an arm that was once optimal but is no longer ideal. To address this challenge, researchers have introduced adaptive techniques, such as variants of UCB that employ a sliding window or exponential discounting (Garivier & Moulines, 2011). These methods proactively forget outdated information by introducing hyperparameters (such as window size or a discount factor), but their performance is consequently tightly coupled with the selection of these parameters, posing a common challenge in the field.

To handle non-stationarity more robustly, the research community has explored more advanced strategies. One direction is meta-algorithms, which adapt to changes by scheduling multiple base learners (Pacchiano et al., 2024). However, such methods often rely on strong assumptions about the base algorithms, and their theoretical guarantees are typically for expected regret, making it difficult to bound worst-case performance in a single run. Another direction involves designing more sophisticated hybrid memory mechanisms within a single policy. Some researchers have proposed a dual-view algorithm within the Thompson Sampling framework that combines a discount factor and a sliding window, attempting to merge the advantages of both mechanisms (Cavenaghi et al., 2021). Despite these valuable contributions to solving non-stationary MAB problems, significant research gaps remain. First, these advanced hybrid memory concepts have not been systematically transferred to and validated within the equally important and widely used UCB family of algorithms. Second, for sliding-window-based algorithms, the impact of a critical design choice—whether to use a "global shared window" or "per-arm independent windows"—lacks an in-depth empirical comparison.

This paper aims to fill the aforementioned research gaps. The core motivation is to optimize and evaluate UCB-family algorithms in non-stationary environments. To this end, this paper first adapts the design philosophy of the dual-view (f-dsw) approach from the Thompson Sampling framework to the UCB framework, proposing and implementing the FDSW-UCB algorithm. This algorithm maintains both a discount-based long-term trajectory and a sliding-window-based short-term trajectory for each arm, integrating the decision-making evidence from both perspectives through an aggregation strategy. Furthermore, the "per-arm independent window" design is implemented and specifically evaluated across a range of adaptive algorithms. For a realistic comparison, a data-driven, semi-synthetic simulation platform was constructed based on two widely used real-world datasets: the MovieLens-1M movie rating data and the Open Bandit Dataset (OBD) (Harper & Konstan, 2015; Saito et al., 2021) e-commerce advertising logs. This approach preserves the complex distributional characteristics of real-world rewards. On this platform, three types of environments—stationary, abrupt, and gradual—are designed to systematically compare the performance of the proposed FDSW-UCB against baseline algorithms such as UCB1, D-UCB, and SW-UCB.

This research finds that D-UCB suffers from a forgetting problem in long-term interactions, leading to linear growth in its cumulative regret. In contrast, SW-UCB using "per-arm independent windows" demonstrates remarkable adaptability by recovering quickly after abrupt environmental changes. The performance of the proposed FDSW-UCB algorithm is highly dependent on the choice of its aggregation strategy. Among the strategies tested, an aggregation-based approach effectively balances long-term and short-term information, exhibiting robust performance across various scenarios.

## 2 METHOD

### 2.1 Experimental Materials: Benchmark Datasets

To ensure that the simulation environment reflects the complexity of real-world problems, two representative and publicly available benchmark datasets are selected.

The MovieLens-1M dataset, released by the GroupLens Research project, is a classical benchmark dataset widely used in the field of recommender systems. It contains approximately one million rating records provided by around 6,000 users for nearly 4,000 movies. Each rating is an integer ranging from 1 to 5. In addition to the ratings, the dataset also provides rich user metadata, including age, gender, and occupation, as well as genre information for the movies. This dataset was selected for two main reasons: first, the multidimensional user features offer a solid foundation for constructing "arms" with intrinsic heterogeneity using clustering methods; second, the five-point rating scale allows to create a reward environment characterized by a non-Bernoulli and multi-modal distribution.

The Open Bandit Dataset is a large-scale, real-world logged bandit feedback dataset publicly released by ZOZOTOWN, a major Japanese online fashion retail platform. This dataset is designed to promote reproducible research on offline evaluation, and it contains interaction logs recorded when various bandit algorithms were used in a real production environment to recommend fashion items to users. In this study, a subset of this dataset that includes 80 distinct items was utilized, each corresponding to a different "arm." The motivation for selecting this dataset lies in its provision of realistic Bernoulli (click/no-click) reward feedback, which is fundamentally different from the reward structure in the MovieLens dataset. This contrast enables to examine the performance of algorithms under heterogeneous reward settings and assess the generalizability of the derived conclusions.

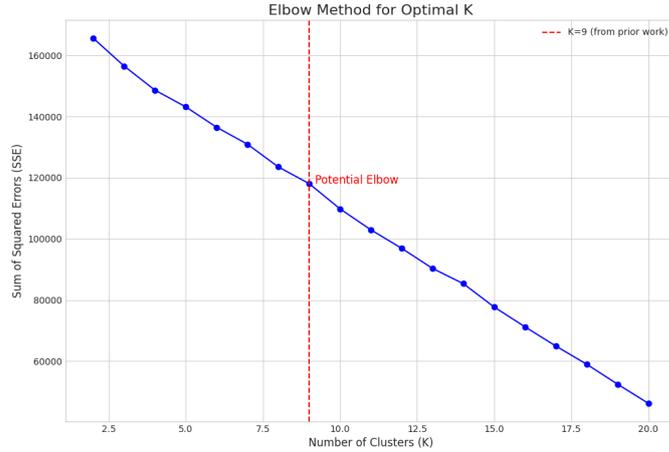

Figure 1: Elbow Method for Optimal K (Movie) (Picture credit : Original).

## 2.2 Construction of a Data-Driven Semi-Synthetic Simulation Environment

The core of this study lies in the design and implementation of a data-driven semi-synthetic simulation platform. This platform is intended to integrate the complexity of real-world data with the flexibility of controlled simulation experiments.

### 2.2.1 Simulation Based on the MovieLens-1M Dataset

In this environment, each "arm" is defined as a user profile group characterised by similar demographic attributes. Specifically, the user features provided in the dataset—namely, age, gender, and occupation—are utilised, and the K-Means clustering algorithm is applied to partition the user population (Ahmed et al., 2020). To guide the selection of the number of clusters, denoted as K, the elbow method is employed. As illustrated in Figure 1, although a distinct "elbow point" is not sharply visible, the curve of the Sum of Squared Errors (SSE) begins to flatten around K=9. This point is considered a reasonable trade-off between model complexity (i.e., the number of clusters) and intra-cluster variance. Consequently, K is set to 9, resulting in nine distinct user clusters, each of which serves as an independent arm in the simulation environment. The reward mechanism for each arm is driven by the actual rating data within the corresponding user cluster. When a multi-armed bandit algorithm selects a specific arm, the simulator randomly samples—with replacement—from all historical movie ratings provided by users in that cluster. The sampled rating, which ranges from 1 to 5 stars, is returned as the reward for that interaction.

### 2.2.2 Simulation Based on the Open Bandit Dataset

The environment based on the Open Bandit Dataset is constructed in a more straightforward manner. The 80 distinct promoted items included in the dataset are directly mapped to the 80 arms within the simulation environment.

Each arm's reward is determined by its associated real-world interaction outcomes. When the algorithm selects a given arm, the simulator performs a random sampling—with replacement—from the historical click records associated with that item. A reward of 1 is returned if the sampled record indicates a click, and 0 otherwise, thereby producing a Bernoulli reward distribution for each arm.

## 2.3 Mechanism for Injecting Non-Stationarity

After constructing the aforementioned stationary environments, a controllable mechanism for injecting non-stationarity is introduced to simulate the temporal dynamics of reward distributions observed in real-world scenarios.

### 2.3.1 Abrupt Drift Mechanism

To model abrupt environmental changes, an abrupt swap drift mechanism **is** designed. When the simulation reaches a predefined time step, the system performs a remapping operation. When the simulation reaches a predefined time step, the system performs a remapping operation. Specifically, this operation instantaneously swaps the underlying reward pools and the precomputed true expected rewards, denoted as $\mu_a$, between the two best-performing arms (Top-2) and the two worst-performing arms (Bottom-2). As shown in Figure 2 and Figure 3, this mechanism is empirically validated: at the moment the drift occurs, the true expected rewards of the affected arms exhibit a precise step change.

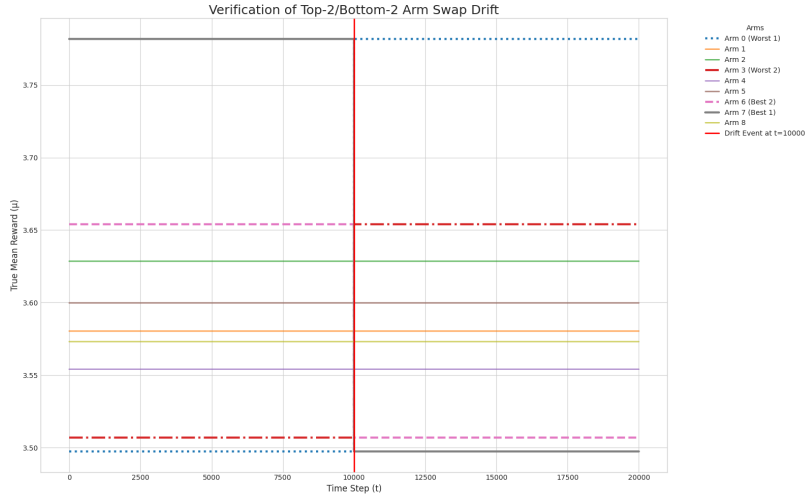

Figure 2: Movie_Abrupt_Drift (Picture credit : Original).

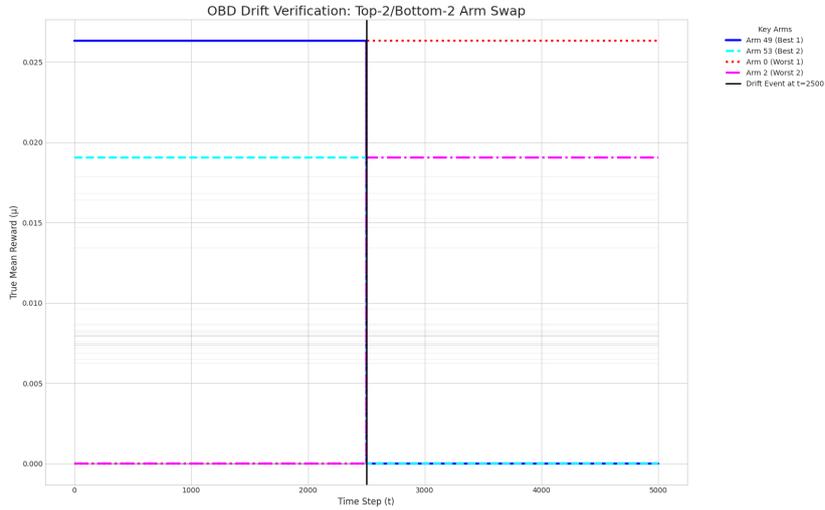

Figure 3: OBD_Abrupt_Drift (Picture credit : Original).

### 2.3.2 Gradual Drift Mechanism

To simulate smoother and more gradual environmental transitions, a gradual drift mechanism is implemented. This mechanism is applied within a predefined time window. This mechanism is applied within a predefined time window. During this period, the true expected rewards, $\mu_a$, of the Top-2 and Bottom-2 arms transition smoothly through linear interpolation. At each time step within this window, the expected reward of each affected arm is adjusted incrementally according to its position on the interpolated trajectory until the complete exchange of values is finalized at the end of the window. As illustrated in Figure 4 and

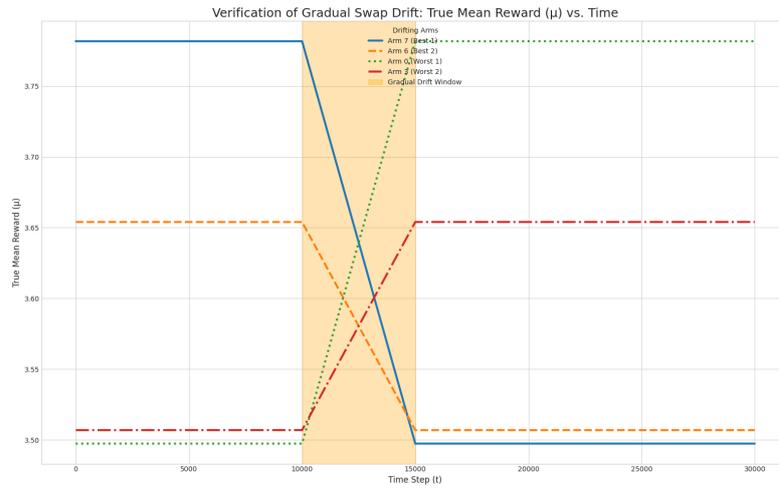

Figure 4: Movie_Gradual_Drift (Picture credit : Original).

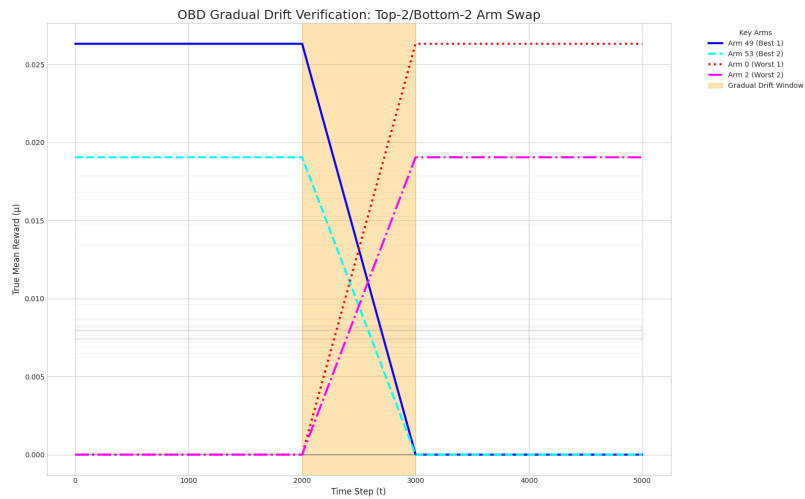

Figure 5: OBD_Gradual_Drift (Picture credit : Original).

Figure 5, the true means of the affected arms exhibit a smooth, linear transformation during the drift interval, accurately simulating progressive environmental dynamics.

### 2.4 Base Algorithms

To establish an effective performance baseline for the subsequent adaptive algorithms, the Upper Confidence Bound 1 (UCB1) algorithm was first implemented. UCB1 is a canonical method for solving the multi-armed bandit problem (Auer et al., 2002), which balances exploration and exploitation through the deterministic principle of "optimism in the face of uncertainty." The core idea of this algorithm is to compute an optimistic upper confidence bound for the value of each arm—the UCB score—and subsequently select the arm with the highest current score.

This UCB score consists of two components. The first is the exploitation term, representing the average observed reward for the arm thus far, which serves as a direct estimate of its true value. The second is the exploration term, which quantifies the uncertainty associated with this value estimate. A key design feature is the introduction of an adjustable positive constant, $\alpha$, which controls the weight of the exploration term. This $\alpha$ value determines the algorithm's propensity for exploration: a higher $\alpha$ amplifies the influence of uncertainty, prompting the algorithm to more aggressively explore less-frequently selected arms, whereas a lower $\alpha$ causes the algorithm to focus more on exploiting known high-reward arms. In a stationary environment, as the number of selections for an arm increases, its exploration term diminishes, enabling the algorithm to eventually converge stably on the optimal arm.

At time step t, the UCB score for each arm a is calculated according to the following rule:

$$\text{UCB1}(a) = \hat{\mu}_a(t-1) + \alpha\sqrt{\frac{\ln(t-1)}{N_a(t-1)}} \quad (1)$$

Where $\hat{\mu}_a(t-1)$ is the observed average reward for arm a up to round t-1, $N_a(t-1)$ is the number of times arm a has been selected up to round t-1, the term t-1 represents the total number of rounds elapsed, and $\alpha$ is a positive constant that controls the level of exploration.

## 2.5 Adaptive Algorithms

To address the challenge of time-varying reward distributions in non-stationary environments, several adaptive algorithms were implemented and evaluated. The core principle of these algorithms is the introduction of a "forgetting" mechanism, enabling them to dynamically adapt to environmental changes. (Garivier & Moulines, 2011).

### 2.5.1 Discounted UCB

The D-UCB (Discounted UCB) algorithm addresses non-stationarity through a mechanism of "gradual forgetting." Unlike standard UCB, which treats all historical data equally, D-UCB introduces a discount factor $\gamma$, where $0 < \gamma < 1$, to exponentially decay past rewards when updating an arm's value estimate. This approach gives more weight to recent rewards in decision-making, while older rewards are gradually "forgotten" due to continuous discounting. This mechanism is widely adopted in research on non-stationarity.

Specifically, instead of maintaining the raw sum of rewards and pull counts, D-UCB maintains its "discounted" versions. At each time step, the cumulative statistics of all arms are multiplied by the discount factor $\gamma$, and only the selected arm is then updated with the new reward and a count of one. As a result, an arm's value estimate is predominantly influenced by its recent performance, allowing the algorithm to track slow changes in the reward means.

At each time step t, the algorithm selects an arm $a_t$ based on the D-UCB score:

$$a_t = a_{a \in A}\left(\frac{S_a^\gamma(t-1)}{N_a^\gamma(t-1)} + \alpha\sqrt{\frac{\ln(\sum_{j \in A} N_j^\gamma(t-1))}{N_a^\gamma(t-1)}}\right) \quad (2)$$

where $S_a^\gamma(t-1)$ is the discounted sum of rewards for arm a and $N_a^\gamma(t-1)$ is its discounted number of pulls, both calculated up to round t-1. These statistics are updated as follows: when arm $a_t$ is selected at time t and receives reward $r_t$, its statistics are updated via $S_{a_t}^\gamma(t) = \gamma S_{a_t}^\gamma(t-1) + r_t$ and $N_{a_t}^\gamma(t) = \gamma N_{a_t}^\gamma(t-1) + 1$. For all unselected arms $j \neq a_t$, their statistics are also decayed: $S_j^\gamma(t) = \gamma S_j^\gamma(t-1)$ and $N_j^\gamma(t) = \gamma N_j^\gamma(t-1)$. The constant $\alpha$ controls the level of exploration.

### 2.5.2 SW-UCB (Sliding-Window UCB)

The SW-UCB algorithm adopts a more direct mechanism of "abrupt forgetting." It estimates an arm's value based solely on recent observations within a sliding window of a fixed size $\tau$, while completely discarding all historical information outside this window. This mechanism theoretically allows SW-UCB to respond more rapidly to abrupt environmental changes, as it can quickly purge the "contamination" from obsolete information.

In this project, a key structural enhancement to the conventional implementation of SW-UCB is introduced by employing a "Per-Arm Window" model. Whereas many traditional implementations use a single, global data window shared across all arms, the design maintains an independent, fixed-size reward queue, $W_a$, for each arm a. This design is more flexible and robust in the common scenario where arms are pulled with non-uniform frequencies, as it prevents new data from a frequently pulled arm from flushing out the valuable, and still relevant, recent information of an infrequently pulled arm.

However, the performance of SW-UCB is highly dependent on the choice of the window size $\tau$, a well-known challenge. If $\tau$ is set too small, the algorithm becomes overly reliant on a few samples, leading to noisy and unstable value estimates. Conversely, if $\tau$ is set too large, the algorithm incorporates too much outdated information, making its response to environmental changes sluggish.

To address this challenge, a heuristic for setting the window size is implemented. This mechanism does not adjust the window size at runtime; instead, it is set once before the experiment begins, based on global parameters. The core idea is to align $\tau$ with the average length of the stationary phases in the environment. Specifically, the size is calculated as follows:

$$\tau = c \cdot \frac{T}{N} \quad (3)$$

where T is the total interaction horizon of the experiment, N is the total number of predefined environmental changepoints, and c is an adjustable constant factor that controls the window's size relative to the average length of a stable period. The intuition is to set the algorithm's "memory" span to approximately cover one full stationary period.

At each time step t, the algorithm selects an arm $a_t$ based on the SW-UCB score:

$$a_t = \underset{a \in \mathcal{A}}{\operatorname{argmax}}(\hat{\mu}_a^{SW}(t-1) + \alpha \sqrt{\frac{\ln(\sum_{j \in \mathcal{A}}|W_j(t-1)|)}{|W_a(t-1)|}}) \quad (4)$$

where $\hat{\mu}_a^{SW}(t-1)$ is the average of rewards within arm a's independent window $W_a(t-1)$, $|W_a(t-1)|$ is the number of samples in that window, $\sum_{j \in \mathcal{A}}|W_j(t-1)|$ is the total number of samples across all arms' windows, and $\alpha$ is the exploration constant.

### 2.5.3 FDSW-UCB (f-Discounted Sliding-Window UCB)

The core of FDSW-UCB lies in constructing a "Dual-View" evaluation system for each arm (Sagi & Rokach, 2018; Cavenaghi et al., 2021). At every decision point, the algorithm concurrently computes two independent UCB scores for each arm a:

The Historical View: A score denoted as $U_a^D$, calculated using the logic of D-UCB, which reflects the arm's long-term, smoothed value.

The Recent View: A score denoted as $U_a^{SW}$, calculated using the logic of SW-UCB, which reflects the arm's short-term, dynamic value.

Subsequently, a configurable aggregation function f fuses these two scores from different temporal dimensions into a single, final decision score. The choice of this aggregation function is therefore critical to the algorithm's performance.

The execution of the algorithm is formalized in the following two steps. First, the component UCB values are computed. Then, an aggregation function f is used to combine these values and select an arm.

The arm $a_t$ selected at time step t is determined by:

$$a_t = \underset{a \in \mathcal{A}}{\operatorname{argmax}}(f(U_a^D(t), U_a^{SW}(t))) \quad (5)$$

where $U_a^D(t)$ is the score for arm a calculated using the D-UCB formula, and $U_a^{SW}(t)$ is the score for arm a calculated using the SW-UCB formula.

To systematically investigate the impact of the aggregation strategy on the algorithm's behavior, the following three aggregation functions for f were implemented and evaluated: Mean

(Balanced Strategy):

$$f(x, y) = (x + y)/2 \quad (6)$$

This strategy gives equal importance to both views, seeking a balanced estimate.

Max (Optimistic Strategy):

$$f(x, y) = max(x, y) \quad (7)$$

This strategy adopts an optimistic stance, assigning a high score to an arm if at least one of the views considers it to be promising.

Min (Pessimistic Strategy):

$$f(x, y) = min(x, y) \quad (8)$$

This strategy adopts a conservative stance, assigning a high score only when both views concur that the arm is performing well.

## 3 EXPERIMENTAL SETUP

To ensure the thoroughness and statistical validity of the experiments, all simulations adhere to the following common configuration:

Interaction Rounds (Horizon): Each independent experimental run continues for a horizon of T = 100,000 time steps. Such a long horizon was chosen to provide algorithms with sufficient time to learn and converge, and to clearly observe their long-term behavior, such as the potential for "catastrophic forgetting" in D-UCB.

Independent Runs: For each combination of experimental scenario and algorithm, N = 3 independent and reproducible simulation runs were conducted. As described in the methodology section, each run was initialized with a different random seed to ensure statistical independence. The final performance curves reported are the average of these three runs, with the standard deviation presented as a shaded region to illustrate the stability of the algorithm's performance.

To comprehensively evaluate the adaptability of the algorithms, an experimental matrix was designed, comprising two datasets and three environmental dynamics, resulting in a total of six distinct testing scenarios:

MovieLens-Stationary: In the MovieLens environment, the reward distributions of the arms remain constant throughout the experiment. This scenario serves to validate the baseline performance of all algorithms on a standard MAB problem.

MovieLens-Abrupt: Four abrupt changepoints are introduced into the stationary MovieLens environment. Specifically, at time steps t = 30,000, t = 45,000, t = 60,000, and t = 90,000, the reward pools of the Top-2 and Bottom-2 arms are swapped to simulate instantaneous, drastic shifts in the environment.

MovieLens-Gradual: Two periods of gradual drift are introduced. The drifts are initiated at t = 30,000 and t = 60,000, each lasting for 10,000 steps. During these intervals, the true means of the Top-2 and Bottom-2 arms undergo a smooth linear exchange.

OBD-Stationary: The reward distributions of the arms in the Open Bandit Dataset environment remain constant.

OBD-Abrupt: Similar to the MovieLens-Abrupt scenario, four abrupt changepoints are introduced into the OBD environment at t = 30,000, t = 45,000, t = 60,000, and t = 90,000.

OBD-Gradual: Similar to the MovieLens-Gradual scenario, two periods of gradual drift are introduced into the OBD environment, initiated at t = 30,000 and t = 60,000.

The following primary metric is employed to measure and analyze algorithm performance:

Primary Metric: Cumulative Expected Regret This is the standard for evaluating the performance of MAB algorithms. The semi-synthetic simulation platform provides access to an oracle perspective, allowing for knowledge of the true expected reward of the optimal arm., $\mu_t^*$, and the true expected reward of the arm actually chosen by the algorithm, $\mu_{a_t}$, at every time step t. Therefore, the expected regret at time step t is defined as:

$$r_t = \mu_t^* - \mu_{a_t} \tag{9}$$

The total cumulative expected regret, R_T, is the sum of the expected regrets over all time steps:

$$R_T = \sum_{t=1}^{T}(\mu_t^* - \mu_{a_t}) \tag{10}$$

## 4 RESULTS AND DISCUSSION

This section presents the performance of the evaluated algorithms across the six experimental scenarios described in Section 3. A quantitative summary of the performance for all experiments, detailing the mean and standard deviation of the final cumulative regret, is compiled in Table 1 and Table 2.

The dynamic cumulative regret curves are, in turn, visualized in Figure 6, Figure 7, Figure 8, Figure 9, Figure 10 and Figure 11.

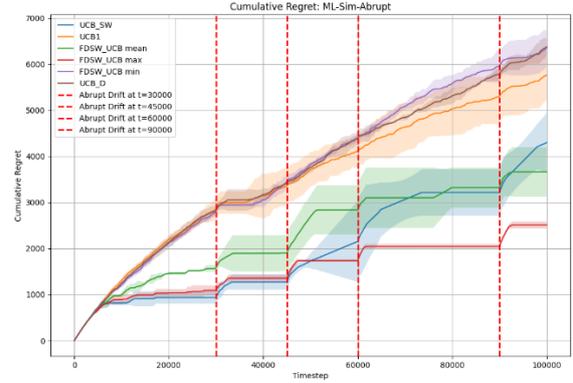

Figure 6: ML-Sim-Abrupt_cumulative_regret (Picture credit : Original).

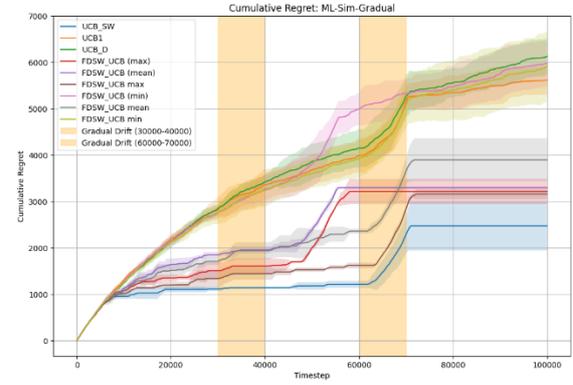

Figure 7: ML-Sim-Gradual_cumulative_regret (Picture credit : Original).

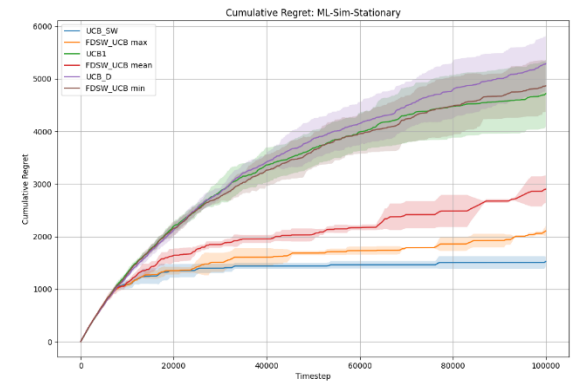

Figure 8: ML-Sim-Stationary_cumulative_regret (Picture credit : Original).

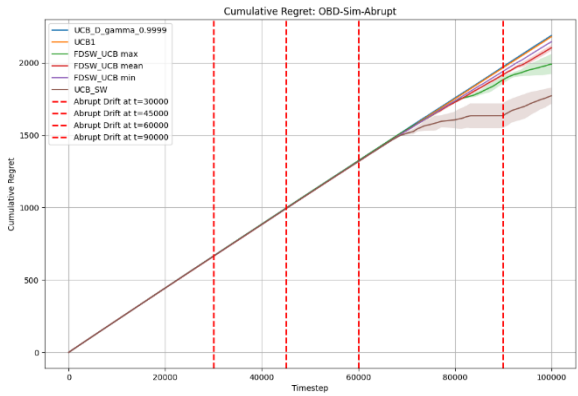

Figure 9: OBD-Sim-Abrupt_cumulative_regret (Picture credit : Original).

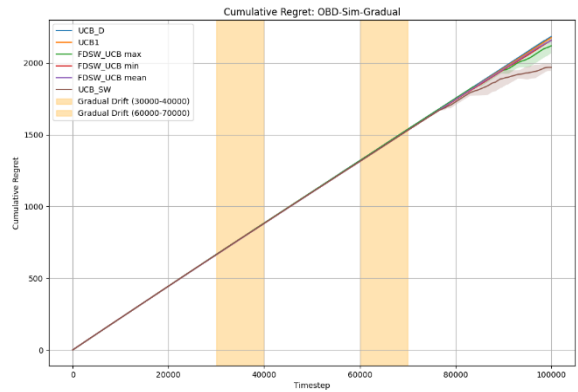

Figure 10: OBD-Sim-Gradual_cumulative_regret (Picture credit : Original).

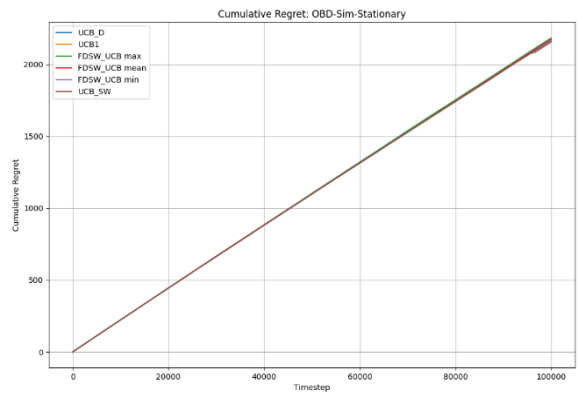

Figure 11: OBD-Sim-Stationary_cumulative_regret (Picture credit : Original).

## 4.1 The Double-Edged Sword of Forgetting Mechanisms

The results of this study reveal the dual role that "forgetting" mechanisms play in non-stationary MAB algorithms, acting as a double-edged sword.

On one hand, the hard-forgetting mechanism of the sliding window (SW-UCB) is shown to be a cornerstone for building robust adaptive algorithms. In Figure 6, the cumulative regret curve of SW-UCB exhibits a distinct "staircase-like" pattern: between environmental changepoints, the curve's growth is flat, indicating effective learning. At the changepoints, the regret increases sharply, but the algorithm quickly adapts to the new environment by purging obsolete information, allowing the regret curve to flatten once more. While it may not be the optimal choice in every scenario (e.g., its performance is inferior to that of FDSW-UCB (max) in ML-Sim-Abrupt), it provides a highly reliable and predictable performance baseline, consistently and significantly outperforming the no-forgetting UCB1.

On the other hand, the gradual-forgetting mechanism of discounting (D-UCB) exposes its catastrophic inherent flaw under the long-horizon setting. The most decisive evidence is found in Figure 11: even in a completely stationary environment, the cumulative regret of D-UCB grows linearly. This profoundly demonstrates the failure of its learning process. The root cause lies in the fact that the discount factor $\gamma$ prevents the exploration term in its UCB formula from ever converging to zero. This forces the algorithm into a state of "compulsory exploration," where it periodically wastes resources exploring sub-optimal arms even after the optimal arm has been identified, leading to continuous, linear regret accumulation. This is not merely a case of "excessive forgetting" but a fundamental failure of the algorithm's learning capability.

## 4.2 The Risks and Opportunistic Nature of Ensemble Strategies

The performance of the designed FDSW-UCB algorithm serves as an excellent case study on the complexities of ensemble algorithm design.

Unexpectedly, in Figure 6, FDSW-UCB (max) emerges as the top-performing algorithm, with its final regret (2511.36) being significantly lower than that of SW-UCB (4299.22). This reveals the opportunistic nature of the max aggregation strategy. The max policy allows the algorithm to leverage the more optimistic estimate from either D-UCB or SW-UCB at any given moment. In the aftermath of an abrupt environmental change, a sub-algorithm might generate a spuriously high UCB value due to noise or a few chances high rewards. The max strategy enables FDSW-UCB to seize this random opportunity, thereby locking onto the new optimal arm faster than any single algorithm could. This is a high-risk, high-reward strategy that capitalizes on system uncertainty.

Table 1: Final Average Cumulative Regret on ML Simulations.

| Algorithm | Stationary | Abrupt | Gradual |
|---|---|---|---|
| UCB1 | 4717.66 ± 627.94 | 5762.55 ± 535.03 | 5611.72 ± 309.47 |
| UCB_D | 5288.85 ± 529.96 | 6377.22 ± 183.61 | 6126.34 ± 391.86 |
| UCB_SW | **1525.00 ± 91.71** | 4299.22 ± 643.57 | **2473.29 ± 531.64** |
| FDSW_UCB (min) | 4866.94 ± 494.98 | 6359.99 ± 382.61 | 5945.52 ± 622.53 |
| FDSW_UCB(mean) | 2902.00 ± 258.88 | 3663.03 ± 536.61 | 3597.29 ± 232.84 |
| FDSW_UCB (max) | 2104.86 ± 97.91 | **2511.36 ± 78.30** | 3186.69 ± 189.60 |

Table 2: Final Average Cumulative Regret on OBD Simulations.

| Algorithm | Stationary | Abrupt | Gradual |
|---|---|---|---|
| UCB1 | 2175.29 ± 8.20 | 2180.37 ± 5.47 | 2177.20 ± 4.21 |
| UCB_D | 2182.90 ± 3.61 | 2188.83 ± 1.80 | 2184.55 ± 7.08 |
| UCB_SW | **2157.13 ± 13.02** | 1772.14 ± 56.49 | **1970.29 ± 25.70** |
| FDSW_UCB (min) | 2165.88 ± 5.17 | 2145.71 ± 7.33 | 2160.21 ± 3.84 |
| FDSW_UCB(mean) | 2175.02 ± 3.64 | 2104.02 ± 19.38 | 2157.54 ± 10.41 |
| FDSW_UCB (max) | 2182.90 ± 3.61 | 1991.01 ± 68.47 | 2120.25 ± 51.12 |

In stark contrast is the thorough failure of the min strategy. This policy requires a high score from both views, a mechanism of pessimistic consensus. Because the D-UCB view suffers from the fundamental flaw of non-converging exploration, the min strategy is almost entirely dragged down by it, inheriting the worst of its behavioral patterns and resulting in performance that is near the bottom in all scenarios.

The core insight from this finding is that a simple ensemble of algorithms does not guarantee a performance improvement; the design of the aggregation function itself is the decisive factor. The aggregation strategy defines the "personality" of the ensemble algorithm: whether it is opportunistic, conservative, or balanced. A truly robust ensemble algorithm may not require a simple max or min, but rather a more intelligent meta-learning mechanism capable of dynamically assessing and weighting the reliability of its different internal information sources.

### 4.3 The "Masking" Effect of Data Characteristics on Performance Disparities

Finally, a comparison of the results from the MovieLens (ML) and Open Bandit Dataset (OBD) reveals a fundamental difference. In the ML environments, the performance curves of the different algorithms are clearly distinct, making the merits and drawbacks of their designs readily apparent. In all OBD scenarios (e.g., Figure 10), however, the regret curves of all algorithms are tightly clustered, rendering their macroscopic performance nearly indistinguishable.

It is posited that this is caused by a difference in the Signal-to-Noise Ratio (SNR). The rewards in the OBD environment are from a Bernoulli distribution of clicks, where the click-through rate (CTR) is typically extremely low. In this setting, the difference between the true reward means of the arms (the "signal") is very faint and is submerged in the massive amount of inherent randomness (the "noise") of the Bernoulli trials. Consequently, the primary bottleneck of the experiment shifts from "how to adapt to environmental changes" to "how to detect a weak signal amidst a sea of noise." This fundamental statistical difficulty overwhelms the sophisticated designs of the various adaptive mechanisms. If an algorithm cannot reliably distinguish the better arm even during a stationary period, then adaptability becomes a moot point.

This finding has significant practical implications: no single algorithm is universally optimal. The choice of algorithm must be matched to the data characteristics of the problem domain. For low-SNR environments, investing resources in complex adaptive mechanisms may yield diminishing returns; the key to the problem may instead lie in developing more efficient exploration methods to accomplish basic signal detection.

### 4.4 Parameter Sensitivity and Research Limitations

While interpreting the findings above, it is imperative to prudently discuss a central theme of this research: hyperparameter sensitivity. The choices of the discount factor $\gamma$ for D-UCB and the window size $\tau$ for SW-UCB are not trivial implementation details but are key factors that profoundly influence, and even dictate, algorithmic behavior. The experiments revealed the "catastrophic forgetting" problem of D-UCB in long-horizon settings. The selection of $\gamma$ places the algorithm in a dilemma:

If $\gamma$ is set very close to 1 (e.g., 0.9999), D-UCB will approximate standard UCB1 over a finite horizon, performing well in stationary environments. However, the trade-off is that its ability to adapt to non-stationarity becomes extremely sluggish because it "forgets" too slowly.

If a smaller $\gamma$ is chosen, the algorithm can forget more quickly, but as demonstrated by the experiments, its exploration term fails to converge, preventing effective learning in any environment, including stationary ones.

Therefore, the performance of D-UCB is highly dependent on the choice of $\gamma$. Finding a $\gamma$ that strikes a balance between "effective learning" and "rapid forgetting" is itself a non-trivial optimization problem, which greatly limits the reliability and ease of use of D-UCB in practice. These findings do not negate the effectiveness of D-UCB in specific (often shorter-horizon) scenarios but rather expose the structural risks inherent in long-horizon applications, driven by its parameter sensitivity.

Similar to D-UCB, the performance of SW-UCB is also highly dependent on the choice of window size $\tau$. In this study, a heuristic was used to calculate the window size. It must be emphasized that this method utilizes global information known before the experiment begins—the total horizon T and the number of changepoints N. To some extent, this provides SW-UCB with an "oracle-like" advantage, as in real-world online applications, it is generally not possible to know in advance when or how many times the environment will change.

Therefore, the superior performance of SW-UCB in these experiments should be understood as the potential upper bound of the algorithm under an ideal parameter configuration. This also constitutes a limitation of the current study. A more rigorous and realistic test would require evaluating the performance degradation of SW-UCB when $\tau$ is set sub-optimally, or exploring more complex algorithms that attempt to tune the window size automatically online. The conclusion is that while the sliding-window mechanism is effective, its efficacy in practice is tightly bound to the challenge of "how to set $\tau$ intelligently."

The discussion of *and tau* also provides a deeper perspective on the behavior of FDSW-UCB. The reason FDSW-UCB (max) could opportunistically succeed in the ML-Abrupt scenario is precisely that it leveraged the "optimal" estimate generated by its two well-configured sub-modules during specific phases. This further suggests that the future research direction for advanced ensemble algorithms should not be limited to exploring static aggregation functions like max or mean. Rather, it should shift towards dynamic, adaptive aggregation strategies—for instance, a meta-algorithm capable of dynamically adjusting its trust in the D-UCB view (which may be more reliable in stable periods) and the SW-UCB view (more reliable during periods of change) based on signals of environmental stability.

## 5 CONCLUSIONS

To address the challenge of accurately evaluating multi-armed bandit algorithms in non-stationary environments, this study designed and implemented a data-driven, semi-synthetic simulation platform. This platform effectively integrates the distributional characteristics of real-world data with the controllability of a simulated environment, thereby enabling a precise measurement of algorithmic adaptability.

The experimental results clearly reveal the inherent properties of two mainstream forgetting mechanisms. The superior performance of a well-configured sliding-window mechanism (SW-UCB) in responding to environmental changes was validated. Concurrently, the catastrophic performance degradation of the discounting mechanism under long horizons, caused by its non-converging exploration, was exposed. Furthermore, experiments with the proposed FDSW-UCB algorithm demonstrated that the performance of an ensemble algorithm is highly dependent on its aggregation strategy.

A limitation of this study is the absence of a comprehensive hyperparameter search. Therefore, future work will systematically investigate the sensitivity to key hyperparameters, such as the window size ($\tau$) and the discount factor ($\gamma$). Additionally, the exploration of more intelligent, dynamically adjustable aggregation strategies will be pursued to design even more robust adaptive algorithms.